\documentclass[10pt,twocolumn,letterpaper]{article}

\usepackage{cvpr}
\usepackage{times}
\usepackage{epsfig}
\usepackage{graphicx}
\usepackage{amsmath}
\usepackage{amssymb}

\usepackage[nolist]{acronym}
\usepackage{booktabs}
\usepackage{subcaption}

\usepackage[breaklinks=true,bookmarks=false]{hyperref}

\cvprfinalcopy 

\def\cvprPaperID{****} 

\begin{document}

\graphicspath{{ARTWORK/}}

\title{Toward Automatic Threat Recognition for Airport X-ray Baggage Screening with Deep Convolutional Object Detection}

\author{
	Kevin~J~Liang$^{1}$\thanks{Correspondence to: \texttt{kevin.liang@duke.edu}},
	John~B.~Sigman$^{1}$,
	Gregory~P.~Spell$^{1}$,
	Dan~Strellis$^{2}$,\\
	William~Chang$^{2}$,
	Felix~Liu$^{2}$,
	Tejas~Mehta$^{2}$,
	and~Lawrence~Carin$^{1}$\\
	$^{1}$Duke University~~ $^{2}$Rapiscan Systems
}

\maketitle

\begin{abstract}
  For the safety of the traveling public, the \acf{TSA} operates security checkpoints at airports in the United States, seeking to keep dangerous items off airplanes.
  At these checkpoints, the \ac{TSA} employs a fleet of X-ray scanners, such as the Rapiscan 620DV, so \acfp{TSO} can inspect the contents of carry-on possessions. However, identifying and locating all potential threats can be a challenging task.
  As a result, the \ac{TSA} has taken a recent interest in deep learning-based automated detection algorithms that can assist \acp{TSO}.
  In a collaboration funded by the \ac{TSA}, we collected a sizable new dataset of X-ray scans with a diverse set of threats in a wide array of contexts, trained several deep convolutional object detection models, and integrated such models into the Rapiscan 620DV, resulting in functional prototypes capable of operating in real time.
  We show performance of our models on held-out evaluation sets, analyze several design parameters, and demonstrate the potential of such systems for automated detection of threats that can be found in airports.
\end{abstract}

\section{Introduction}
The \acf{TSA} oversees the safety of the traveling public in the United States of America. 
One of the most visible functions of the \ac{TSA} is security screening of travelers and their personal belongings for potential threats.
Handsearching each passenger's bag would be both time-consuming and intrusive, so X-ray scanner systems such as the Rapiscan 620DV are deployed to remotely provide an interior view of baggage contents.
Many real threats are captured nationwide: in 2018, for example, 4239 firearms were found in carry-on bags, and more than 80\% of these were loaded~\cite{air_traffic_by_the_numbers}.
These numbers have steadily grown in recent years as air traffic has continued to increase nationally.
The capability of finding these objects effectively is an important concern for national security.

Currently, the detection of prohibited items relies on \acfp{TSO} to visually pick out these items from displayed image scans.
This is challenging for several reasons.
First, the set of prohibited items that \acp{TSO} must identify is quite diverse: firearms; sharp instruments; blunt weapons; and \acp{LAG} with volumes exceeding the \ac{TSA}-established thresholds all pose security concerns.
Second, the majority of scans are benign, yet \acp{TSO} must remain alert for long periods of time.
Third, because X-ray scans are transmission images, the contents of a bag appear stacked on top of each other into a single, often cluttered scene, which can render identification of individual items difficult.
The Rapiscan 620DV provides dual perpendicular views to ameliorate this problem, but depending on the orientations, views can still be non-informative.
Finally, given the need to maintain passenger throughput, evaluation of a particular scan should not take excessively long.

For the aforementioned reasons, an automatic threat detection algorithm to aid human operators in locating prohibited items would be useful for the \ac{TSA}, especially if it can be readily integrated into the existing fleet of deployed scanners.
Fundamentally, the \acp{TSO} both localize and identify dangerous items in an image, which are the same objectives of \textit{object detection}~\cite{girshick2014rich, girshick15fastrcnn, ren2015faster, liu2016ssd, Dai2016, huang2017speed}. 
Object detection has long been considered a challenging task for computers, but advances in deep learning~\cite{Goodfellow2016} in recent years have resulted in enormous progress.
Specifically, \acp{CNN}~\cite{lecun1989backpropagation} have proven extremely useful at extracting learned features for a wide variety of computer vision tasks, including object detection.
As a result, the \ac{TSA} is interested in assessing the feasibility of deploying algorithms that can automatically highlight objects of interest to \acp{TSO}~\cite{TSA_kaggle_competition}.

\begin{figure*}[t]
	\centering
	\includegraphics[width=.975\linewidth]{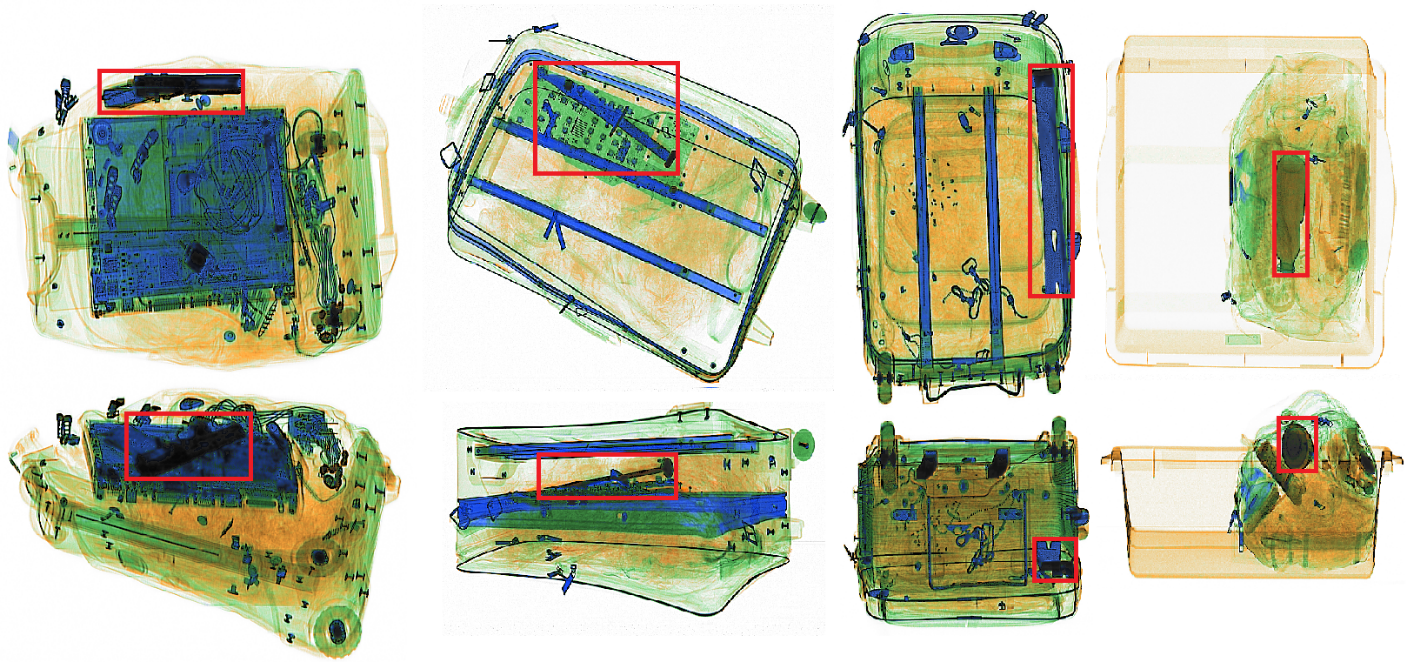}
	\caption{Example scans of bags in false color containing a firearm (handgun), sharp (knife), blunt (crow bar), and \ac{LAG} (bottle of liquid), from the left to right, with (top row) top view and (bottom row) side view shown. Ground truth locations of threats in each image are bounded with a red box. Scans produced by a laboratory prototype not in TSA configuration.}
	\label{fig:ex_threats}
\end{figure*}

Most deep learning methods require a large training dataset of labeled examples to achieve good performance~\cite{Sun2017}; for object detection, this means data comprising both images and bounding boxes with class labels.
While many such datasets exist for \ac{RGB} natural scenes (\textit{e.g.}~\cite{everingham2010pascal, lin2014microsoft, Cordts2016}), none contain threats in X-ray luggage, and so a sizable data collection effort was necessary for this endeavor.
We assembled a large variety of cluttered bags (\textit{e.g.} clothing, electronics, etc.) with hidden threats (firearms, sharps, blunts, \acp{LAG}), and scanned these with the Rapiscan 620DV.
Each threat in the scans was then annotated with a tight bounding box and labeled according to class.
This dataset was then used for training and evaluating object detection models.

In this work, we present the results of a research effort in collaboration with the \ac{TSA} to develop a deep learning-based automated threat detection system. 
We first describe the Rapiscan 620DV scanner and the data collection process.
We then introduce the deep learning algorithms we used to perform object detection and how we integrated them into a Rapiscan 620DV prototype, for live testing.
Finally, we present experimental results on a number of models we tested on the collected data.
The resulting prototype system has shown great promise, and technology like this may one day be deployed by the \ac{TSA} to airports nationally.

\section{Data Collection}
\subsection{Rapiscan 620DV X-Ray Scanning System}
The Rapiscan 620DV X-ray screening system is designed for aviation and high-security applications. 
It comprises a tunnel 640 mm wide and 430 mm high, equipped with a 160 kV / 1 mA X-ray source that achieves a steel penetration of 33mm and a wire resolution of about 80 micrometers (40 American Wire Gauge). 
The scanner produces two views through the near-orthogonal orientation of the fan-shaped beams from the X-ray sources. 
These projections generate a horizontal and vertical view of the object under inspection, both of which can be used to identify the contents of a bag.
X-ray detectors collect both high and low X-ray energy data, which allows for material discrimination. Examples are shown in Figure \ref{fig:ex_threats}. 

While it is possible to use the high and low energy image scans as direct inputs to our model, we instead choose to use the pre-processed \ac{RGB} coloration typically shown to human \acp{TSO}.
This coloring uses the relationship between the linear attenuation coefficient and photon energy to estimate effective atomic number ($Z$), transforming the image into one where material properties can be more readily inferred: for example, organic materials tend to have low $Z$, while metallic materials tend to have higher $Z$.
According to Rapiscan's proprietary coloring scheme, metallic objects are colored blue, organic materials are tinted orange, and materials with effective $Z$ ($Z_{eff}$) between these two are shaded green.
Using this false coloring as our input achieves two objectives: $(i)$ encoding of additional human knowledge of material properties, which are highly informative for threat detection (firearms, sharps, and blunts, for example, often contain metallic components) and $(ii)$ aligning the image input color distribution closer to the pre-trained weights, which were trained on \ac{RGB} natural scenes.

\subsection{Scan Collection and Annotation}
Baggage scans were collected at various sites, occurring over multiple collection events.
This data collection targeted several of the \ac{TSA}'s designated threat categories: firearms (\textit{e.g.} pistols), sharps (\textit{e.g.} knives), blunts (\textit{e.g.} hammers), and \acp{LAG} (\textit{e.g.} liquid-filled bottles).
A diverse set of unique items from each class were selected to provide coverage for each threat type; for example, the firearms set included both assembled and disassembled guns.
To simulate the diversity of real-world traffic, a variety of host bags was used, including roller, laptop, and duffel bags.
Each was filled with diverse assortments of benign items, such as clothing, shoes, electronics, hygiene products, and paper products.
Threats were added to each host bag in different locations and orientations, as well as with imaginative concealments, to simulate the actions of potentially malicious actors.
Under the assumption that threat objects are typically rare, most bags contained only one threat, as in the examples shown in Figure \ref{fig:ex_threats}.

Given the time-consuming nature of assembling bags for scanning, a single bag was used to host different unique threats for multiple scans, with a minor exchanging of benign clutter between insertions.
Each bag was also scanned in several different poses (\textit{e.g.} flipped or rotated).
These strategies allow for more efficient collection of more scans and encourage our models to learn invariance to exact positioning within the tunnel.
Total number of threats scanned are summarized in Table \ref{tab:threat_count}.

\begin{table}[t]
	\centering
	\resizebox{1.\columnwidth}{!}{%
		\begin{tabular}{ c || c | c }
			\toprule
			Threat Type & Total Threats & Total Images \\
			\midrule
			Blunts & 10 & 3366 \\
			Firearms & 43 (assembled) + 19 (disassembled) & 3480 \\
			\acp{LAG} & 70 & 3456 \\
			Sharps & 40 & 3484 \\
			\bottomrule
		\end{tabular}
	}
	\caption{Total number of unique threat items and number of images collected for each threat.} \label{tab:threat_count}
\end{table}
After the scans were collected, each image was hand-annotated by human labelers, where each label consisted of both the threat class-type, as well as the coordinates of the bounding box.
Each box was specified to be as tight as possible in each view, while still containing the full object; in the case of objects like sharps and blunts, this definition included the handle, for instances in which there was one.
In total, the entire data collection effort of assembling, scanning, and labeling bags took over 400 worker hours.

\section{Methods}
\label{sec:methods}
\subsection{Convolutional Neural Networks}
The advent of deep convolutional neural networks (CNNs)~\cite{lecun1989backpropagation} has resulted in a quantum leap in the field of computer vision.
Across virtually all computer vision tasks, the incorporation of \acp{CNN} into model designs has resulted in significant performance gains; consequently, \acp{CNN} play a significant role in almost every recent computer vision algorithm.
Unlike classical methods that rely upon carefully selected, human-engineered features, machine learning (and deep learning) methods learn these features from the data itself.
\acp{CNN} in particular are designed to learn hierarchical representations~\cite{Zeiler2014}, resulting in a feature extractor that produces highly informative, abstract encodings that can be used for downstream tasks, such as classification~\cite{krizhevsky2012imagenet}.
Additionally, the learned visual features are highly transferable: for example, \ac{CNN} weights learned for the classification task of ImageNet~\cite{Deng2009} can serve as a good initialization for other datasets or even other related computer vision tasks~\cite{Yosinski2014, Razavian2014, girshick2014rich}.
Doing so can considerably reduce the number of training examples needed for the desired task.
In the setting of automatic threat detection at TSA checkpoints, this is especially significant, as we must assemble, scan, and label each training sample ourselves; pre-trained networks allow us to significantly cut down man-hours and costs.

There are several design considerations for \acp{CNN}. 
Most obvious is model performance: how good are the features the \ac{CNN} extracts for the downstream task?
In general, there is a positive correlation between the number of \ac{CNN} layers (depth) and parameters with overall performance~\cite{Simonyan2015, he2016deep}, though architectural choices can play a significant role as well~\cite{Zoph2018}.
However, finite hardware memory and processing time limit model size.
We consider several popular \ac{CNN} architectures in our experiments, summarized in Table \ref{tab:CNN_sizes}.

\begin{table}[t]
	\centering
	\resizebox{\columnwidth}{!}{%
		\begin{tabular}{ c || c | c }
			\toprule
			\ac{CNN} Architecture & Top-1 Accuracy & Number of parameters \\
			\midrule
			Inception V2~\cite{inception_v2} & 73.9 & 10.2 M \\ 
			\hline
			ResNet-101~\cite{he2016deep} & 77.0 & 42.6 M\\  
			\hline
			ResNet-152~\cite{he2016deep} & 77.8 & 58.1 M \\  
			\hline
			Inception ResNet V2~\cite{Szegedy2016} & 80.4 & 54.3 M \\ 
			\bottomrule
		\end{tabular}
	}
	\caption{ImageNet classification accuracy and number of parameters for each of the \acp{CNN} architectures considered in our experiments. Adapted from ~\cite{huang2017speed}.} \label{tab:CNN_sizes}
\end{table}

\begin{figure*}[t]
	\centering
	\includegraphics[width=.95\textwidth]{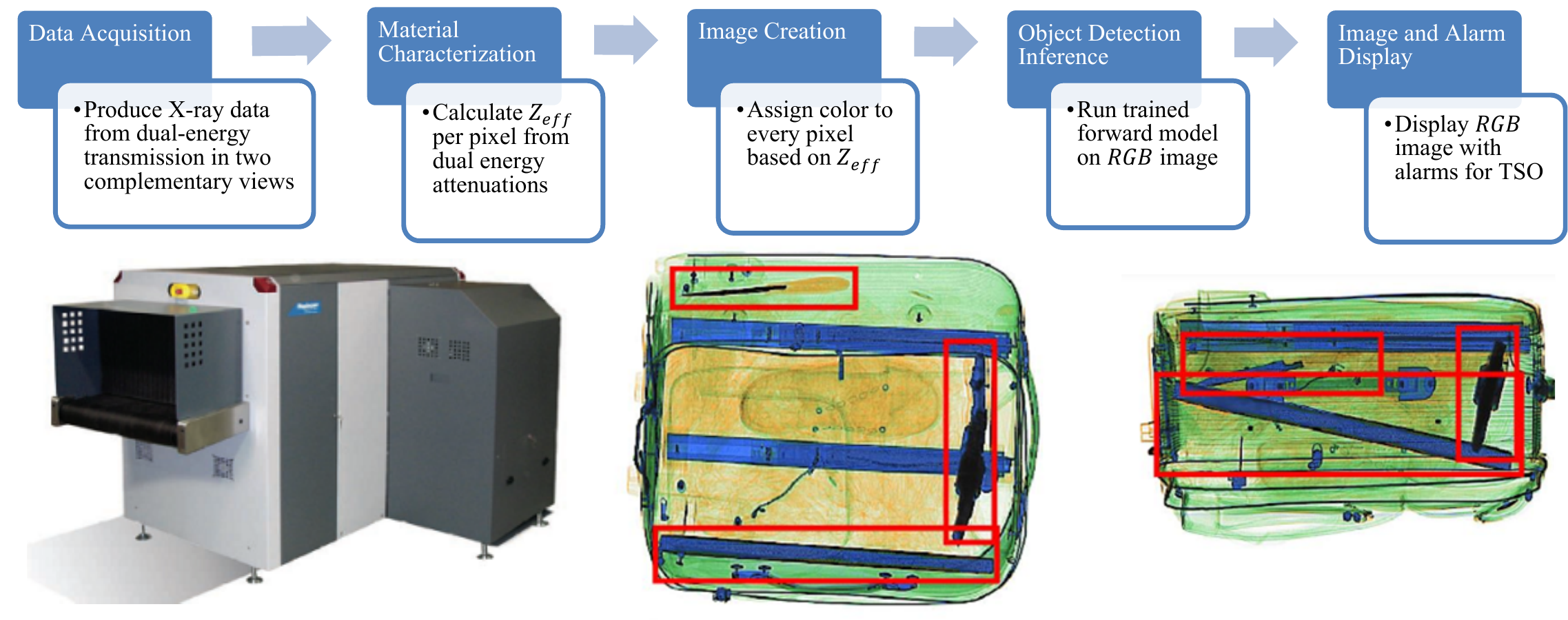}
	\caption{Diagram of the prototype Rapiscan 620DV X-ray screening system with threat recognition capability. Dual-energy X-Ray information yields false-color images of two views, which are fed to a trained deep convolutional object detector. Detections above threshold are displayed for the user. Scans produced by a laboratory prototype not in TSA configuration.}
	\label{fig:system-flow-chart}
\end{figure*}

\subsection{Object Detection} \label{sec:obj_det}
Localizing and classifying objects in a scene is a canonical research area in computer vision.
In this context, localization refers to the production of a \textit{bounding box} which is as tight as possible while still containing the entire object, while classification is the identification of which of a pre-determined set of classes the object belongs to.
Formally, given an image $X$, the goal of object detection is to predict the class $c_i$ of each object indexed by $i$, as well as the center and dimensions $(x_i, y_i, w_i, h_i)$ of a bounding box.

Modern object detectors are almost exclusively built upon \ac{CNN} backbones.
The specific \ac{CNN} architecture used is often readily interchangeable, with the choice of \ac{CNN} depending on the trade-off between accuracy with speed and memory.
How predictions are made from the features extracted by the \ac{CNN} can vary, and various object detection \textit{meta-architectures}~\cite{huang2017speed} have been recently proposed, of which we highlight two notable ones here.

\textbf{Faster R-CNN}: Faster R-CNN~\cite{ren2015faster} makes predictions in a two-stage process. In first stage, called the \ac{RPN}, a set of reference boxes of various sizes and dimensions (termed \textit{anchor boxes}) are tiled over the entire image. 
Using features extracted by a \ac{CNN}, the RPN assigns an ``objectness'' score to each anchor based on how much it overlaps with a ground-truth object, as well as a proposal of how each anchor box should be adjusted to better bound the object.
The $N_p$ box proposals with highest objectness scores are then passed to the second stage, where $N_p$ is a hyperparameter controlling the number of proposals. 
In the second stage, a classifier and box refinement regressor yield final output predictions. 
Non-maximal suppression reduces duplicate detections.

\textbf{\ac{SSD}}: Unlike Faster R-CNN, which performs classification and bounding box regression twice, Single-stage detectors like \ac{SSD}~\cite{liu2016ssd} combine both stages.
This eliminates the proposal stage to directly predict both classes and bounding boxes at once.
This reduction tends to make the network much faster, though sometimes at the cost of accuracy.

\subsection{Evaluation}
The two-part nature of the object detection task--localization and classification--requires evaluation metrics that assess both aspects of detections.
The quality of an algorithm-produced predicted box ($B_p$) with a ground-truth bounding box ($B_{gt}$) is formalized as the \ac{IoU} $ = \mathrm{area}(B_p \cap B_{gt})/\mathrm{area}(B_p \cup B_{gt})$.

A \ac{$T_p$}, \ac{$F_p$}, and \ac{$F_n$} are defined in terms of the \ac{IoU} of a predicted box with a ground-truth box, as well as the class prediction. 
A true positive proposal is a correctly classified box that has an \ac{IoU} above a set threshold (\textit{e.g.} 0.5), a false positive proposal either misclassifies an object or does not achieve a sufficiently high \ac{IoU}, and a false negative is a ground-truth object that was not properly bounded (with respect to \ac{IoU}) and correctly classified. 

At a particular \ac{IoU} threshold, the \textit{precision} and \textit{recall} of the model may be computed as the proportion of proposed bounding boxes that are correct and the proportion of ground truth objects that are correctly detected, respectively. 
These quantities are: $\mathrm{Precision} = T_p/(T_p + F_p)$, $\mathrm{Recall} = T_p/(T_p + F_n)$.
\ac{PR} curves are constructed by plotting both quantities over a range of operating point thresholds.
We present these curves in Section \ref{sec:experiments} to provide a sense for model performance.
Additionally, we may quantitatively summarize model performance through \ac{mAP}.
\ac{AP} is the \ac{AUC} of the \ac{PR} curve for a single class, and \ac{mAP} is the mean of the \acp{AP} across all classes. 

\begin{table*}[t]
	\centering
	\resizebox{2\columnwidth}{!}{
		\begin{tabular}{ c || c || c || c | c | c | c }
			\toprule
			Model & Speed (s/scan) & \acs{mAP} & Sharps & Blunts & Firearms & \acp{LAG}\\
			\midrule
			SSD-InceptionV2 & 0.042 & 0.7523 & 0.408 & 0.918 & 0.757 & 0.907  \\
			Faster-RCNN-ResNet101 & 0.222 & 0.9166 & 0.766 & 0.976 & 0.944 & 0.973  \\
			Faster-RCNN-ResNet152 & 0.254 & 0.9244 & 0.786 & 0.980 & 0.947 & 0.976 \\
			Faster-RCNN-InceptionResNetV2 & 0.812 & 0.9410 & 0.818 & 0.983 & 0.962 & 0.985  \\
			\bottomrule
		\end{tabular}
	}
	\caption{Inference speed and \ac{mAP} of the considered feature extractor and meta-architecture combinations. Timing measured on a Nvidia GeForce GTX 1080 \ac{GPU}.} 
	\label{tab:model_results}
\end{table*}

\subsection{Rapiscan 620DV Integration}
In order to take a concrete step towards the \ac{TSA}'s goal of potentially deploying the deep learning-based automated threat detector, we also worked to integrate the algorithm with the Rapiscan 620DV.
The Rapiscan 620DV has an onboard computer and monitors to construct and display images from the output of the X-ray photon detectors, as well as algorithms for explosives detection.
We wish to leave these functionalities untouched, simply overlaying an additional detection output on screen.
Therefore, we pipe the constructed scan images to our model, perform inference, and project the predictions to the display (see Figure \ref{fig:system-flow-chart}).

To achieve threat recognition, we export a trained model and run it in parallel with existing software.
The system computer hardware was upgraded to an Intel i7 CPU and a Nvidia GeForce GTX 1080 \ac{GPU} in order to support the TensorFlow~\cite{Abadi2015TF} implementation of the model graph.
This allows for a single integrated machine to perform all of the computation for the 620DV, unlike previous implementations that require an additional auxiliary machine to perform the deep neural network computation \cite{Liang2018}.
While the resulting integrated system has been used for live demos, the experimental results we report in this paper were computed with a held-out test set.

\newcommand{\ModelComparisonFigureWidth}{.495\linewidth}
\begin{figure*}[t]
	\centering
	\begin{subfigure}[b]{\ModelComparisonFigureWidth}
		\includegraphics[width=\linewidth]{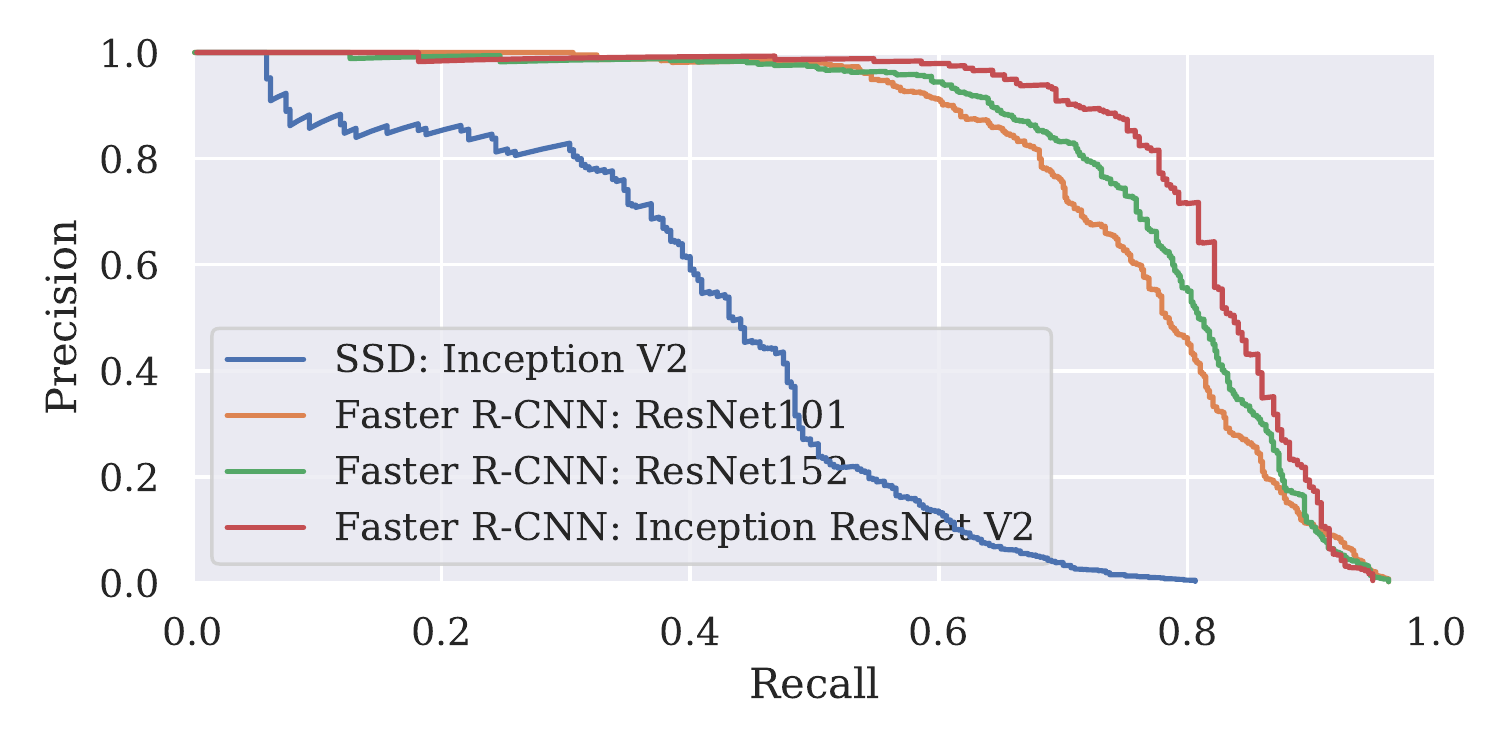}
		\caption{}
		\label{subfig:model-comparison-knives}
	\end{subfigure}
	\begin{subfigure}[b]{\ModelComparisonFigureWidth}
		\includegraphics[width=\linewidth]{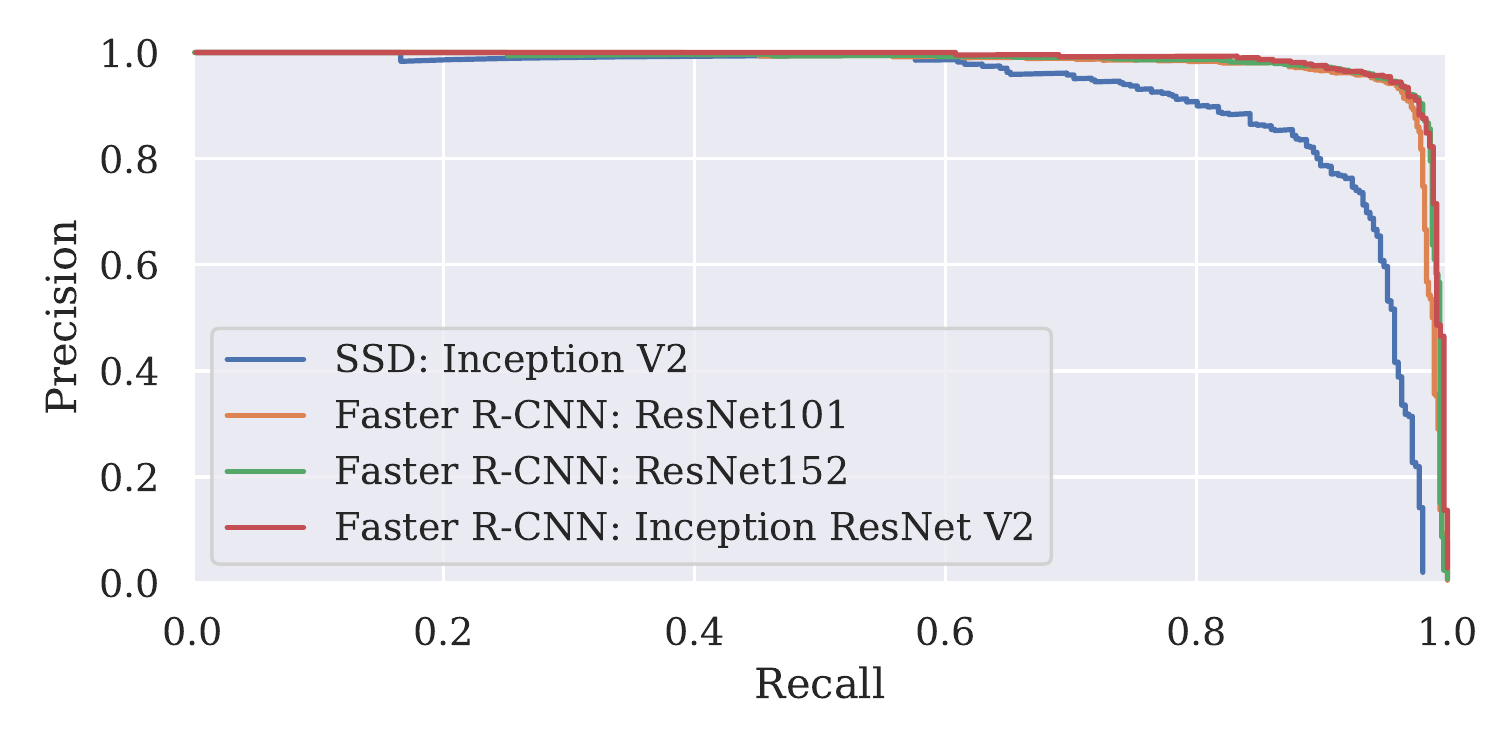}
		\caption{}
		\label{subfig:model-comparison-blunts}
	\end{subfigure}
	\vspace{.3cm}
	\begin{subfigure}[b]{\ModelComparisonFigureWidth}
		\includegraphics[width=\linewidth]{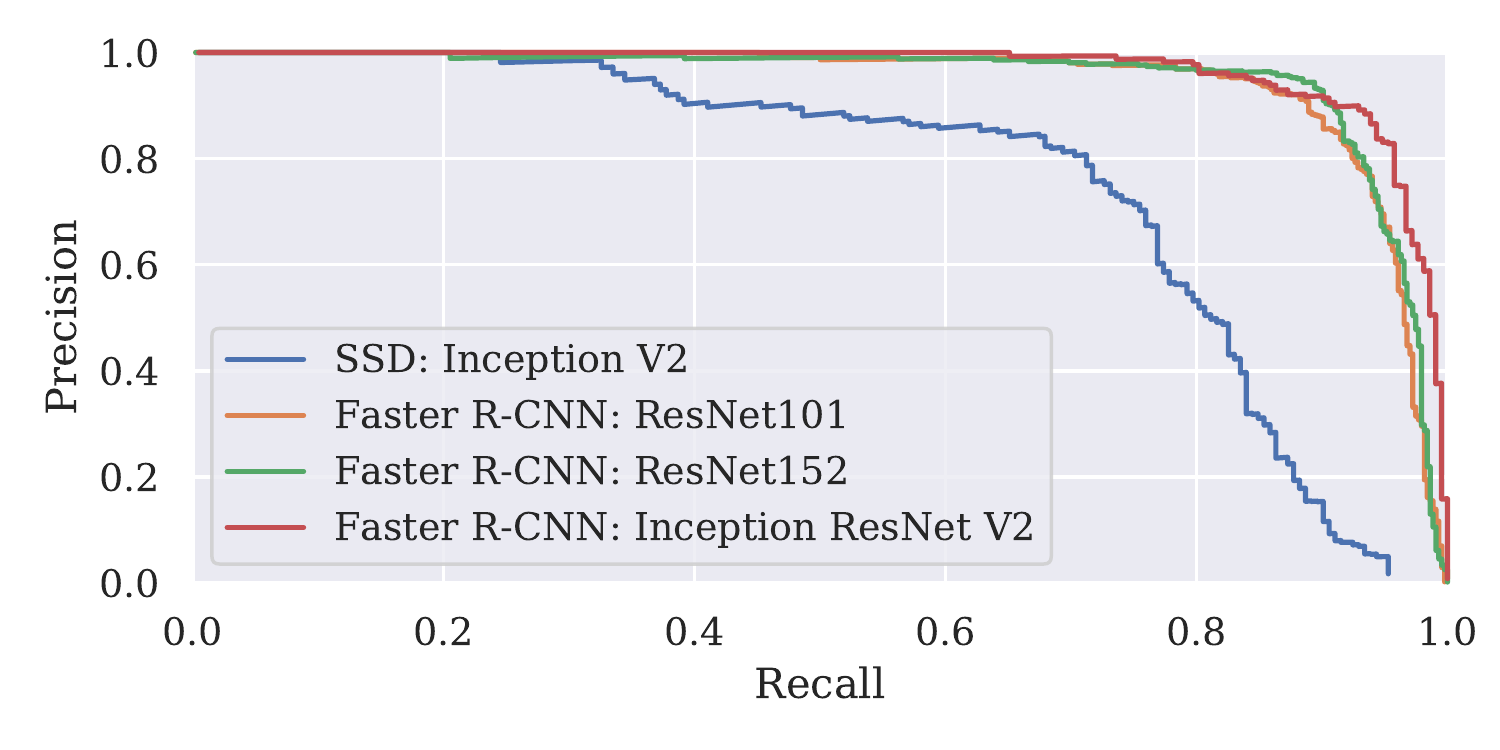}
		\caption{}
		\label{subfig:model-comparison-guns}
	\end{subfigure}
	\begin{subfigure}[b]{\ModelComparisonFigureWidth}
		\includegraphics[width=\linewidth]{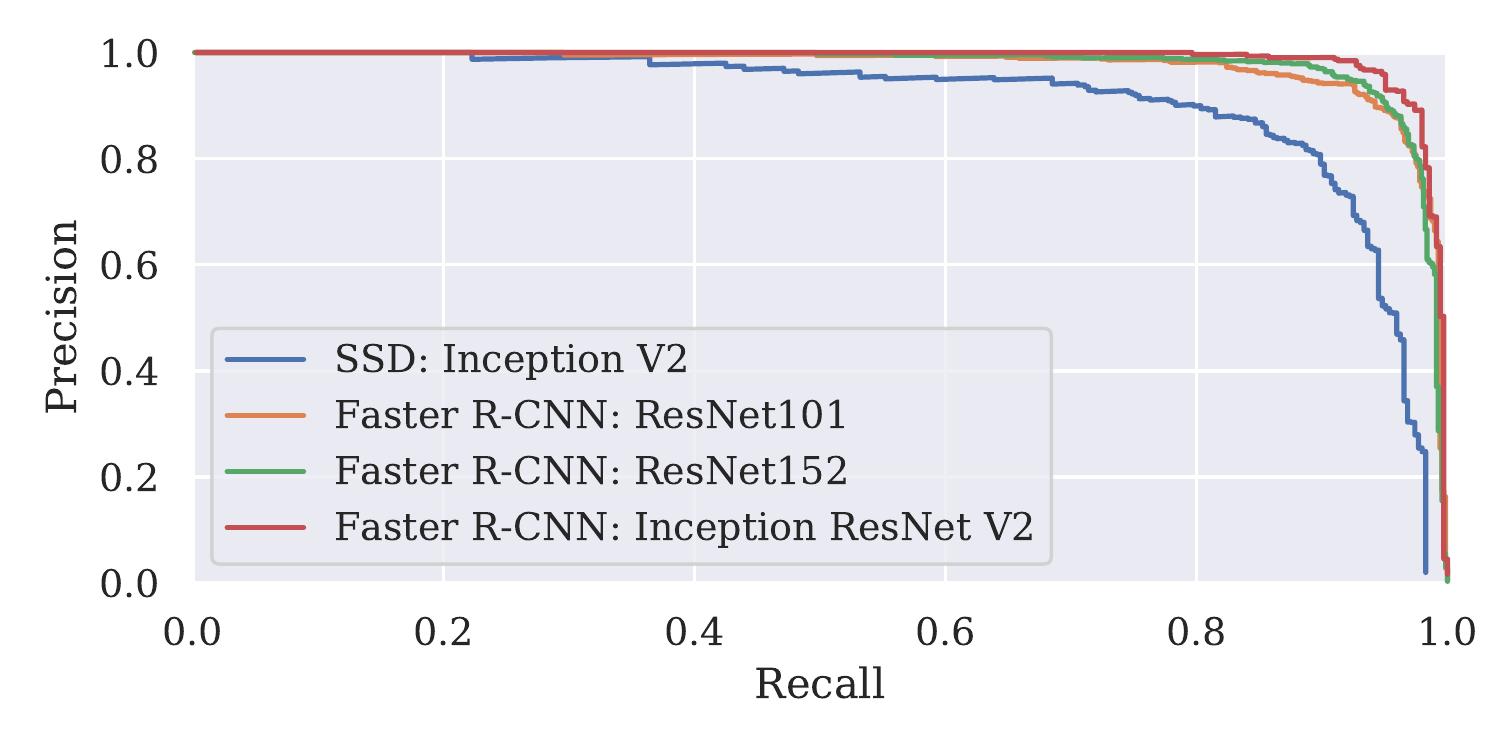}
		\caption{}
		\label{subfig:model-comparison-lags}
	\end{subfigure}
	\caption{\ac{PR} curves for four meta-architecture/feature extractor combinations. (\subref{subfig:model-comparison-knives}) Sharps, (\subref{subfig:model-comparison-blunts}) Blunts (\subref{subfig:model-comparison-guns}) Firearms (\subref{subfig:model-comparison-lags}) \acp{LAG}.}
	\label{fig:pr_curves}
\end{figure*}

\section{Related Work}
The development of computer-aided screening for aviation security has garnered much attention in the past two decades. We focus here specifically on efforts to locate and classify potential threats in X-ray images.

Initial work using machine learning to classify objects in X-ray images leveraged hand-crafted features fed to a traditional classifier such as a Support Vector Machine (SVM).  In particular, \cite{Bastan2011} used Bag-of-Visual-Words (BoVW) and an SVM to classify X-ray baggage with feature representations such as Difference of Gaussians (DoG) in conjuction with scale-invariant feature transform (SIFT) \cite{lowe1999sift}. Further BoVW approaches are used for classification in \cite{Turcsany2013}, \cite{Bastan2013}, \cite{Mery2016}, \cite{Kundegorski2016}.

While deep learning has been applied to general image analysis for at least a decade, its adoption for X-ray security image analysis is relatively recent. 
Still, there are several works that apply deep learning to baggage screening.
In~\cite{rogers2017automated}, the authors provide a review of methods for automating X-ray image analysis for cargo and baggage security, pointing to the use of \acp{CNN} as a promising direction.
The first application of deep learning to an X-ray baggage screening context was for classifying manually cropped regions of X-ray baggage images that contained different classes of firearms and knives, with additional benign classes of camera and laptop~\cite{akcay2016transfer}.
To perform classification, ~\cite{akcay2016transfer} fine-tuned a pre-trained \ac{CNN} to their unique datasets, leveraging transfer learning to improve training with a limited number of images compared to the size of datasets that \acp{CNN} are typically trained on.
In \cite{akcay2016transfer}, the authors compare their classification performance to the BoVW methods mentioned above. 

The work of~\cite{akcay2016transfer} is extended in~\cite{akcay2017evaluation, akcay2018using} to examine the use of deep object detection algorithms for X-ray baggage scans.  The authors address two related problems: binary identification of objects as firearms or not and a multiclass problem using the same classes as \cite{akcay2016transfer}.  They expand the CNN classification architectures investigated to include VGG \cite{Simonyan2015} and ResNet \cite{he2016deep}, and they further adapt Faster R-CNN \cite{ren2015faster}, R-FCN \cite{Dai2016}, and YOLOv2 \cite{redmon2017yolo9000} as CNN-based detection methods to X-ray baggage. 
However, these experiments were done in simulation on pre-collected datasets, without any integration into the scanner hardware. 
They also do not take advantage of the X-ray scanner's multiple views.

Concurrent with this work, the \ac{TSA} has sought to incorporate deep learning systems at U.S. airport security checkpoints in other efforts. 
In~\cite{Liang2018}, the authors present data collection efforts for firearms and sharps classes and compare the performance of five object detection models.
Relative to~\cite{Liang2018}, we also include blunt weapons and \acp{LAG} categories, and we train a single four-class detector, rather than training an individual detector for each category.

\section{Experiments}\label{sec:experiments}
\begin{figure*}[t]
	\centering
	\begin{subfigure}[b]{.475\linewidth}
		\includegraphics[width=\linewidth]{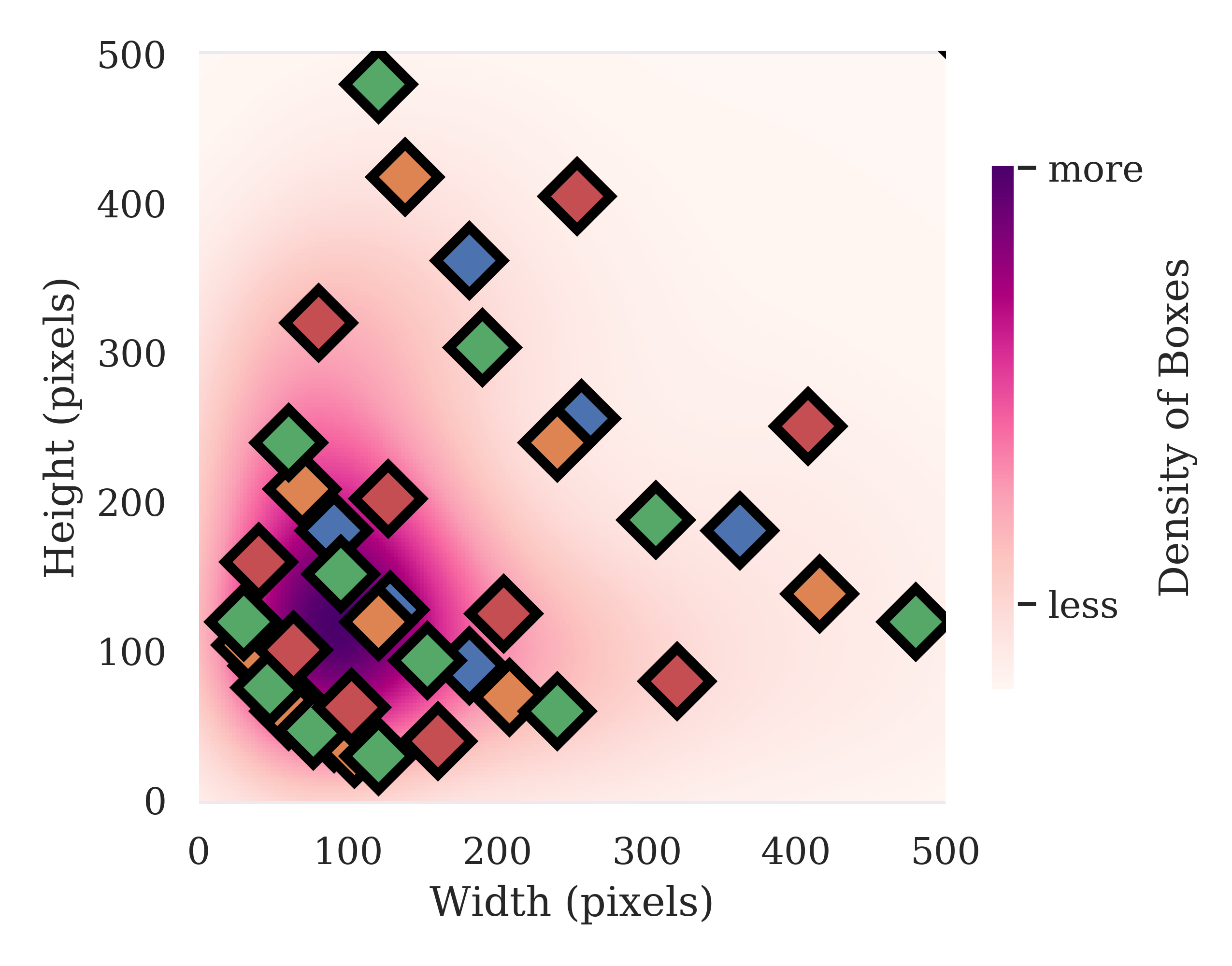}
		\caption{}
		\label{subfig:anchor-density}
	\end{subfigure}
	\begin{subfigure}[b]{.49\linewidth}
		\includegraphics[width=\linewidth]{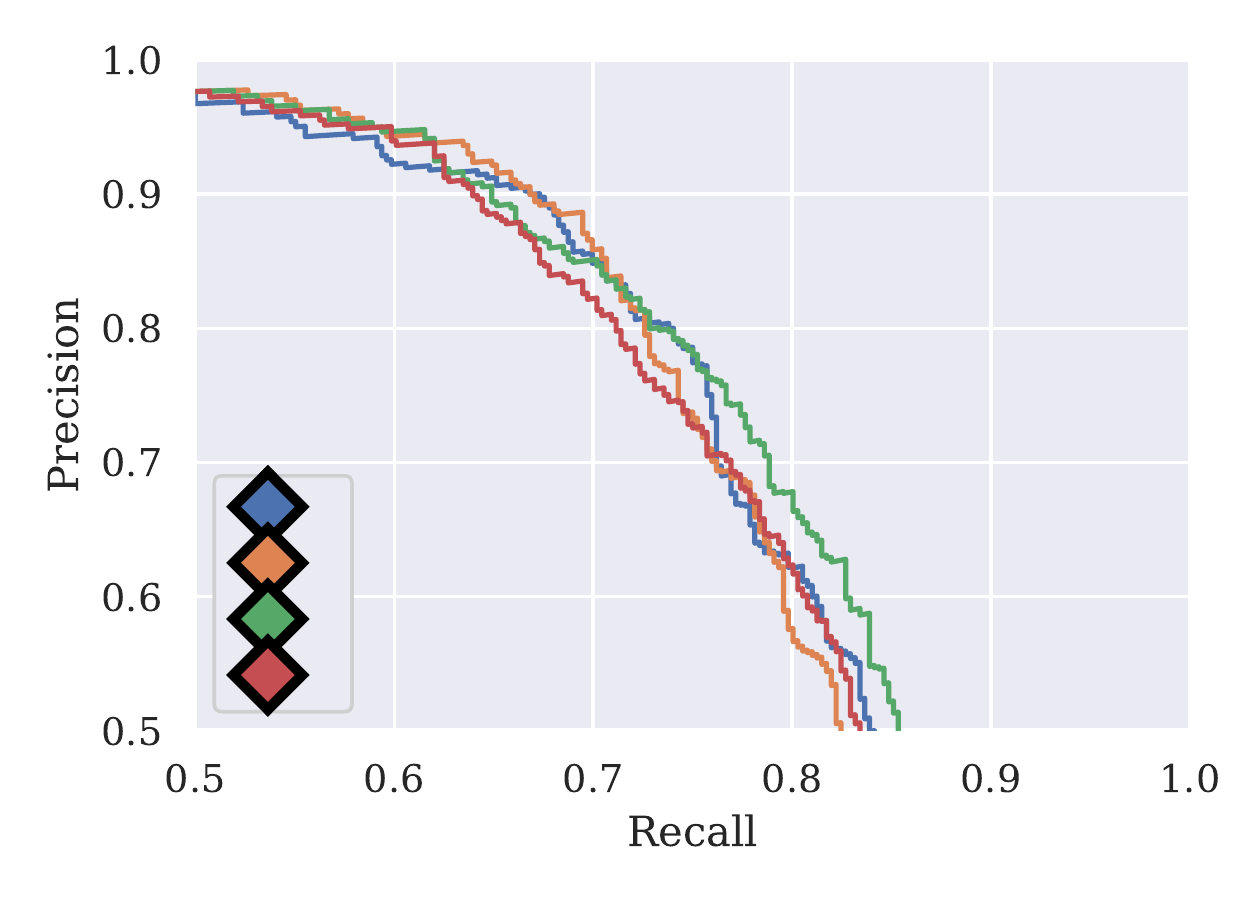}
		\caption{}
		\label{subfig:anchor-results}
	\end{subfigure}
	\caption{(\subref{subfig:anchor-density}) Heatmap indicating density of bounding box dimensions for the training set. Various anchor box distributions are shown; each color indicates a different anchor box experimental setting. Natural image defaults in blue. (\subref{subfig:anchor-results}) \acl{PR} curves for default anchor boxes and engineered set on sharps. Colors correspond to the distributions shown in (\subref{subfig:anchor-density}).}
	\label{fig:anchor_boxes}
\end{figure*}

\begin{figure*}[t]
	\centering
	\begin{subfigure}[b]{.25\linewidth}
		\includegraphics[width=\linewidth]{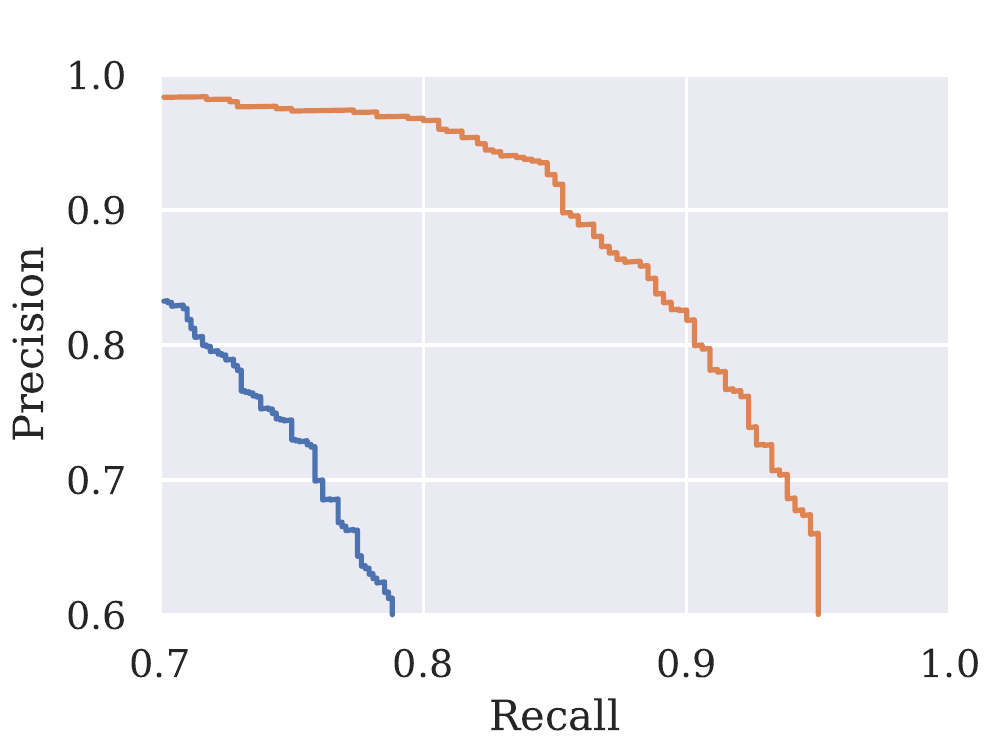}
		\caption{}
		\label{subfig:sharps-multiview-pr}
	\end{subfigure}
	\hspace{-.01\textwidth}
	\begin{subfigure}[b]{.25\linewidth}
		\includegraphics[width=\linewidth]{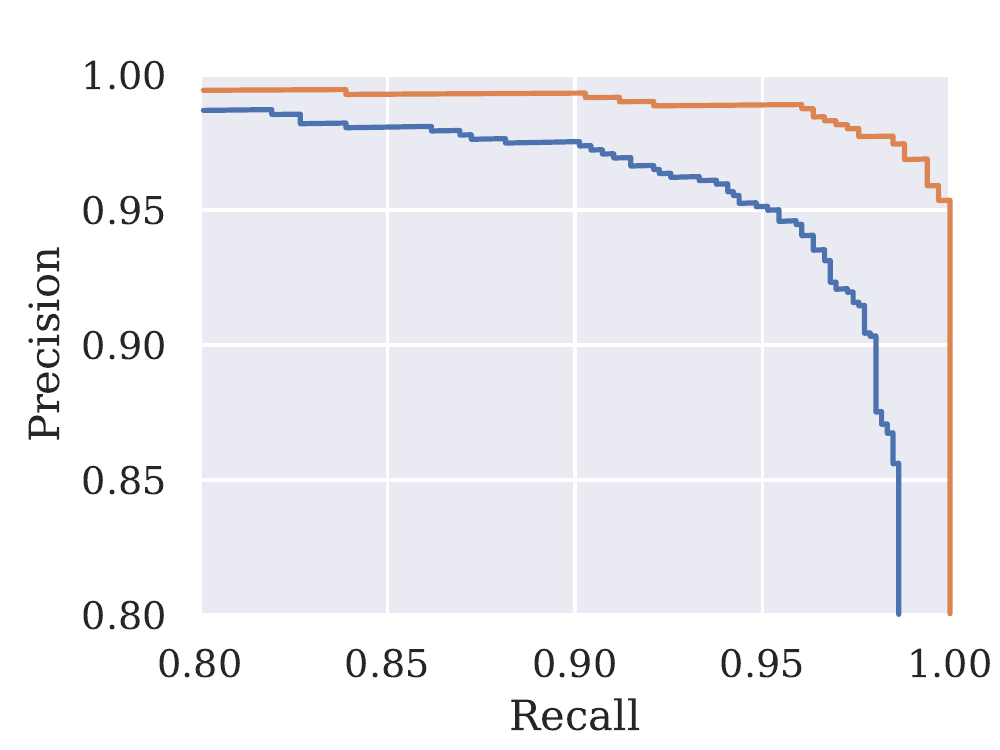}
		\caption{}
		\label{subfig:blunts-multiview-pr}
	\end{subfigure}
	\hspace{-.02\textwidth}
	\begin{subfigure}[b]{.25\linewidth}
		\includegraphics[width=\linewidth]{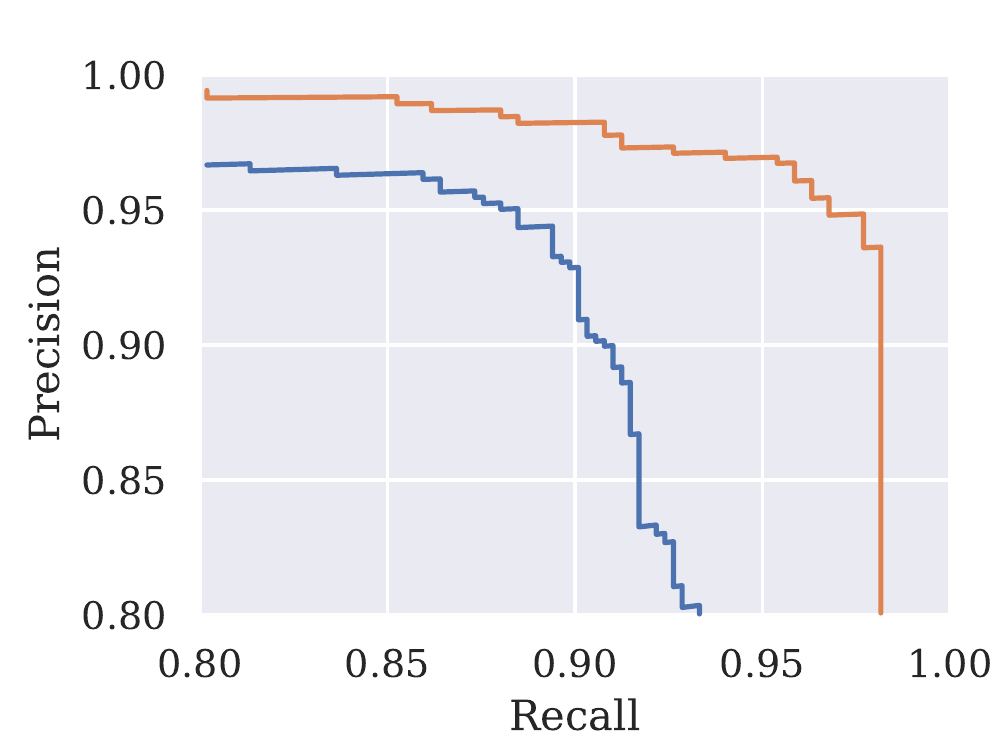}
		\caption{}
		\label{subfig:guns-multiview-pr}
	\end{subfigure}
	\hspace{-.02\textwidth}
	\begin{subfigure}[b]{.25\linewidth}
		\includegraphics[width=\linewidth]{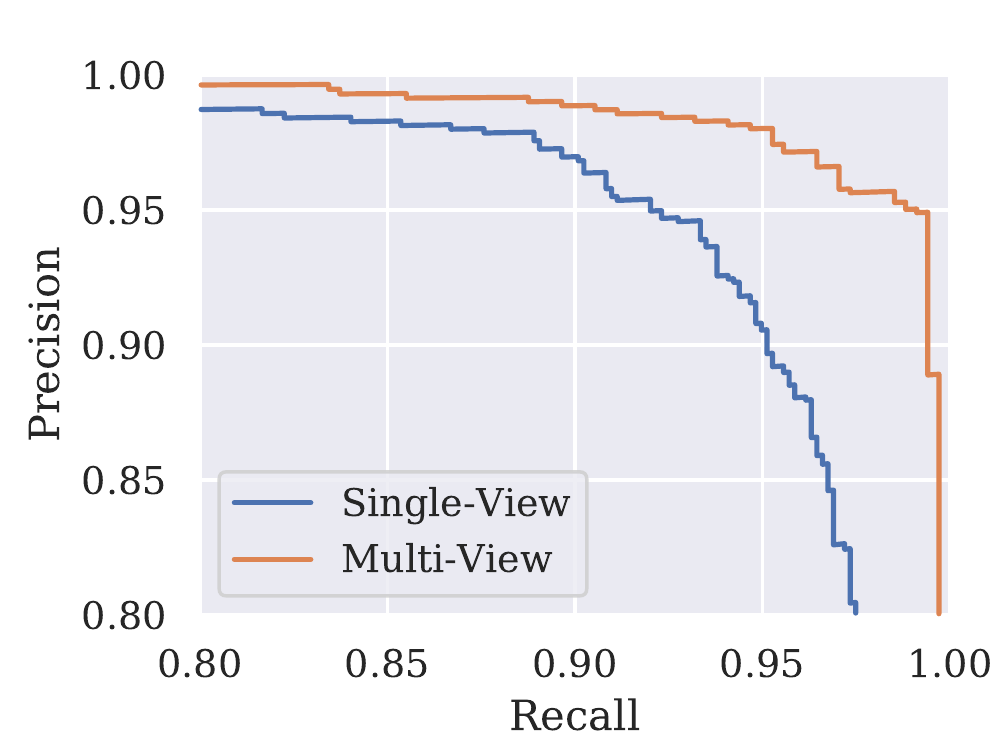}
		\caption{}
		\label{subfig:lags-multiview-pr}
	\end{subfigure}
	\caption{\acl{PR} comparison of single view (blue) versus with multi-view (orange) detection for (\subref{subfig:sharps-multiview-pr}) sharps (\subref{subfig:guns-multiview-pr}) blunts (\subref{subfig:guns-multiview-pr}) firearms (\subref{subfig:lags-multiview-pr}) \acp{LAG} . Note that the multi-view graphs shown here are a choice of analysis, and not a different technique. The training and inference of Faster R-CNN are the same in both traces.}
	\label{fig:multi-view-precision-recalls}
\end{figure*}

For our experiments and in-system implementation, we use Google's code base~\cite{huang2017speed} of object detection models, implemented in TensorFlow~\cite{Abadi2015TF}.
We initialize each model with pre-trained weights from the MSCOCO Object Detection Challenge \cite{lin2014microsoft} and then fine-tune them to detect each of the target classes  (firearms, sharps, blunts, \acp{LAG}) simultaneously, which allows us to perform detection four times as fast as if we trained a separate algorithm for each. Since we initialize with weights pre-trained on MSCOCO, we pre-process each image by subtracting from each pixel the channel-means of the MSCOCO dataset; this aligns our pre-processing with that performed on images for the MSCOCO Challenge. 

For all Faster R-CNN algorithms, we use a momentum optimizer~\cite{momentum1999} with a learning rate of 0.003 for 130,000 steps, reducing it by a factor of 10 for 40,000 steps, and reducing by another factor of 10 for a final 30,000 steps.
For the \ac{SSD} model, we used 200,000 steps of an RMSprop optimizer~\cite{RMSProp2012} with an exponential decay learning rate starting at 0.003, and decaying by 0.9 every 4000 steps.
During training, a batch size of 1 was used for all Faster R-CNN models, and a batch size of 24 was used for SSD.

From the $13,786$ images collected, we create a $70/10/20$ train-validation-test split, which we use for all experiments.
We take care to ensure the two images (views) of a particular bag remain in the same split.

\subsection{Feature Extractor and Meta-architecture}\label{sec:feature-extractors}
As discussed in Section \ref{sec:methods}, there are many options for both \ac{CNN} feature extractor and the object detection meta-architecture, each with its own advantages and disadvantages.
See~\cite{huang2017speed} for extensive comparisons on MS COCO~\cite{lin2014microsoft}.
For the collected X-ray scan dataset, we choose to analyze several high-performing combinations. Detection performance is measured in terms of \ac{AP} for each of the classes of interest, and \ac{mAP} for overall performance is also calculated.
We also measure processing time per scan to project practical passenger wait times.

We summarize the results in Table \ref{tab:model_results} and Figure \ref{fig:pr_curves}.
Overall, Faster R-CNN with Inception ResNet V2 has the highest \ac{mAP}, while SSD with Inception V2 performed the worst.
In general, faster models are less accurate, which may be seen in the ``Speed" column of Table \ref{tab:model_results}.
Faster R-CNN with the two smaller feature extractors (ResNet101 and ResNet152) achieve nearly the same performance on sharps as ResNet Inception V2, but at more than three times the speed.
While the speed of single-stage models is suitable for video frame rates, we found this to be unnecessary for checkpoint threat recognition and to sacrifice too much accuracy.

\subsection{Anchor Boxes}
As discussed in Section \ref{sec:obj_det}, bounding box predictions are typically made relative to anchor boxes tiled over the image.
The object detection algorithms we have considered were primarily designed for finding common objects (\textit{e.g.} people, animals, vehicles) in natural scenes, with datasets like PASCAL VOC~\cite{everingham2010pascal} or MS COCO~\cite{lin2014microsoft} in mind.

The anchor box distribution is commonly held to act as a kind of ``prior'' over the training data.
In YOLO V2~\cite{redmon2017yolo9000}, anchors are learned by k-means clustering, and some of the performance gains of this model are credited to this improvement. 
We chose several configurations of anchor boxes to better match the distribution of our training data, and display those configurations alongside training dataset bounding box dimension density in Figure \ref{subfig:anchor-density}.
The dataset used for these experiments was smaller than the dataset used for the main findings as described in Table \ref{tab:threat_count}.
Training, test, and validation sets were drawn from a pool of images containing 2768 Sharps, 1788 \acp{LAG}, 1800 Blunts, and 3080 Firearms.
This does not impact our conclusions stated in the next paragraph.

During model validation, some of these configurations showed modest gains for sharps, but these did not generalize during testing.
The sharps class \ac{PR} curves for the anchor box distributions in \ref{subfig:anchor-density} are shown in \ref{subfig:anchor-results}.
We find that performance is robust to different anchor configurations, showing that even with a different box size distribution, Faster R-CNN is able to learn accurate bounding box regressors.

\begin{table}[]
	\centering
	\resizebox{\columnwidth}{!}{
		\begin{tabular}{ c || c | c }
			\toprule
			Threat & Single View \ac{AP} & Multiple Views \ac{AP} \\
			\hline
			Sharps & 0.786 & 0.935 \\
			Blunts & 0.980 & 0.995 \\
			Firearms & 0.947 & 0.984 \\
			LAGs & 0.976 & 0.994 \\
			\bottomrule
		\end{tabular}
	}	
	\caption{Single View vs. Effective Multiple View \acp{AP}.}
	\label{tab:multiview_results}
\end{table}

\newcommand{\DetectionSubfigWidth}{.32\textwidth}
\newcommand{\DetectionScale}{.33}
\newcommand{\DetectionAngle}{-90}

\begin{figure*}[t]
	\centering
	\begin{subfigure}[b]{\DetectionSubfigWidth}
		\centering
		\includegraphics[scale=\DetectionScale]{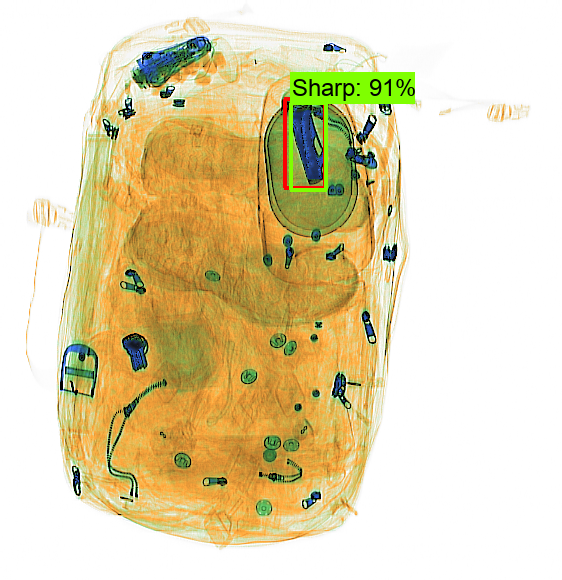}
		\caption{}
		\label{fig:knife-detection}
	\end{subfigure}
	\begin{subfigure}[b]{\DetectionSubfigWidth}
		\includegraphics[scale=\DetectionScale]{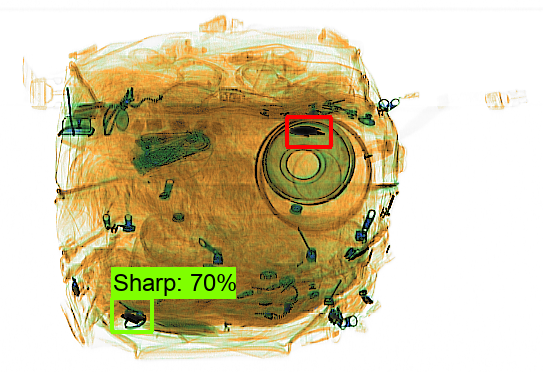}
		\caption{}
		\label{fig:missed-knife-detection}
	\end{subfigure}
	\begin{subfigure}[b]{\DetectionSubfigWidth}
		\includegraphics[scale=\DetectionScale]{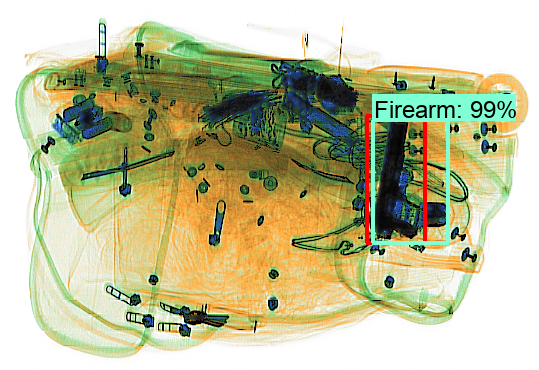}
		\caption{}
		\label{fig:gun-detection}
	\end{subfigure}\\
	\begin{subfigure}[b]{\DetectionSubfigWidth}
		\includegraphics[scale=0.275]{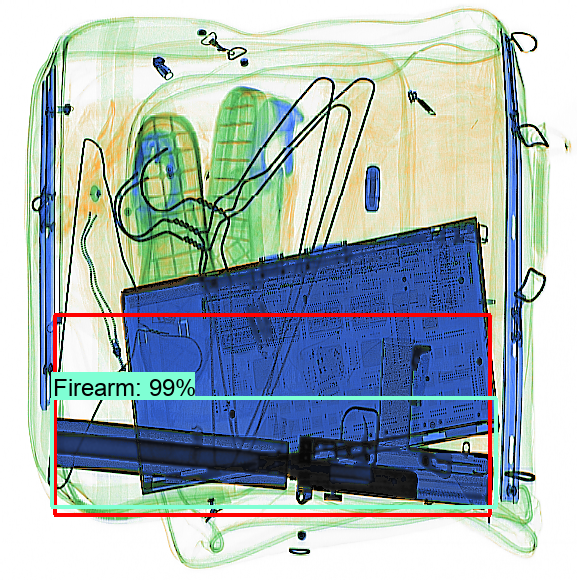}
		\caption{}
		\label{fig:gun-2-detection}
	\end{subfigure}
	\begin{subfigure}[b]{\DetectionSubfigWidth}
		\centering
		\includegraphics[scale=\DetectionScale]{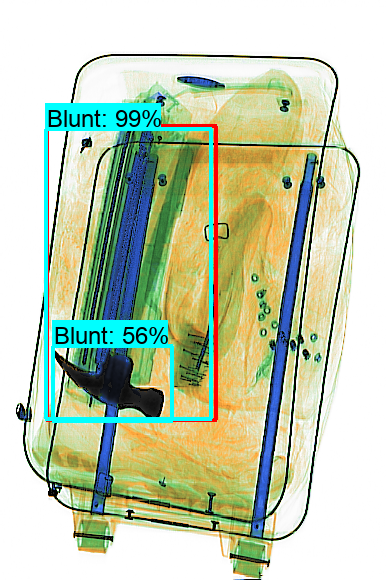}
		\caption{}
		\label{fig:blunt-detection}
	\end{subfigure}
	\begin{subfigure}[b]{\DetectionSubfigWidth}
		\centering
		\includegraphics[angle=-90,scale=\DetectionScale]{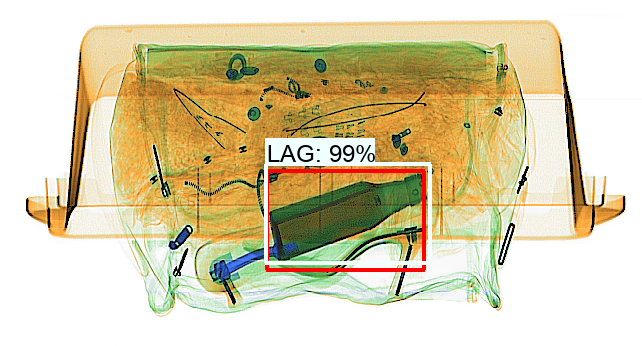}
		\caption{}
		\label{fig:lag-detection}
	\end{subfigure}  
	\caption{Sample detections from the Faster R-CNN model with ResNet152 as the feature extractor. Ground truth bounding boxes are shown in red. (\subref{fig:knife-detection}-\subref{fig:missed-knife-detection}) Sharp detection. (\subref{fig:gun-detection}-\subref{fig:gun-2-detection}) Firearm detection. (\subref{fig:blunt-detection}) Blunt weapon detection. (\subref{fig:lag-detection}) \ac{LAG} detection. The color of predicted box and the label indicate the predicted class. Scans produced by a laboratory prototype not in TSA configuration.}
	\label{fig:sample_dets}
\end{figure*}

\section{Discussion}
The results we have shown bear implications for a pilot real-world deployment of this technology. 
In Table \ref{tab:model_results}, we showed test \ac{AP} on sharps and timing for four feature extractor/meta-architecture pairs.
In a possible real-world system, we strive for inference rates which would not impact screening time and security checkpoint throughput.
Because of the long evaluation time of the Faster R-CNN with InceptionV2 model ($\sim800$ ms seconds per bag), we recommend use of Faster R-CNN with ResNet152 ($\sim250$ ms per bag) for its performance/speed tradeoff.
For the remainder of the Discussion section, we will show results only from this model.

\subsection{Multiple View Redundancy}
\label{sec:multi_view}
Unlike typical object detection research benchmarks, the Rapiscan 620DV provides two views along nearly perpendicular axes of the same scanned object.
Within the context of threat detection in X-ray images, this is especially important, as individual views may occasionally be uninformative due to perspective or clutter.
Leveraging the two separate views can improve overall performance.

In order to account for the multiple views, we consider a true positive in any view to be a true positive in all views.
False positives are added independently across all views.
Note that this is not describing a change to the training of the algorithm, nor the inference process.
Rather, by performing our analysis in this way, we hope to better represent how the system might work in a potential real-world deployment, when both views are available to a \ac{TSO}. 
We show the improvements in \ac{PR} between single-view and multi-view evaluation in Figure \ref{fig:multi-view-precision-recalls} and summarize the \ac{AP} in Table \ref{tab:multiview_results}. 

\subsection{Sample Detections}
In Figure \ref{fig:sample_dets}, we display selected detections from the fully trained Faster R-CNN with ResNet152 as the feature extractor, for a number of threat classes.

In Subfigures (\ref{fig:knife-detection}-\ref{fig:missed-knife-detection}), we display two views of the same scan.
The very small profile of the folding knife in one view  (\ref{fig:missed-knife-detection}) makes detection challenging for the trained object detector (though there is a low-confidence false alarm).
However, the knife presents a more clear profile in Subfigure \ref{fig:knife-detection}, and is detected there.
This motivates what we call ``multi-view'' analysis, which we discuss further in Section \ref{sec:multi_view}.

Subfigure \ref{fig:blunt-detection} shows a blunt threat which is detected twice.
The larger detection, which encompasses the head and handle of a hammer, is a \acl{$T_p$}, because the \ac{IoU} of this detection is greater than 0.5.
The other detection in this image, however, only covers the hammer's head. While the presence of the hammer merits an alarm, the detection does not overlap enough with the ground truth, and is therefore a \acl{$F_p$}.
Some of the training data included hammer heads disconnected from a handle.
It may be harder for the \ac{CNN} to learn to bound hammers with handles or hammer heads only.

To demonstrate detections of the remaining threat classes Subfigure \ref{fig:lag-detection} shows a detected \ac{LAG}, and Subfigures \ref{fig:gun-detection} and \ref{fig:gun-2-detection} show scans with firearms.
Note that the machine pistol in Subfigure \ref{fig:gun-2-detection} is not as well localized, compared to the firearm in \ref{fig:gun-detection}, likely due to the obscuring presence of a laptop. 
However, such an alarm still makes the threat readily visible to a human operator.

\section{Conclusion}
We have investigated use of state-of-the-art techniques for the challenging task of threat detection in bags at airport security checkpoints.
First, we collected a significant amount of data, assembling by hand many bags and bins which simulate everyday traffic.
These concealed a wide variety of threats.
We scanned each bag to produce X-ray images, and annotated both views of the scan.
We then trained multiple modern object detection algorithms on the collected data, exploring a number of settings and engineering them for the task at hand.
We have presented the results of evaluating the model on held-out validation and test data.

In general, we do not find single stage methods to be accurate enough as a security screening method, and their frame rate advantages are superfluous in this application.
There are variants of the Faster R-CNN which can run on commercially available computer hardware, and still achieve accurate threat recognition.

In addition to the evaluation presented in Section \ref{sec:experiments}, the \ac{TSA} has also tested prototype Rapiscan 620DV systems with directly integrated trained models.
These results have shown the promise of deep learning methods for automatic threat recognition.
Further, they illustrate that the \ac{TSA}, using X-ray scanners such as the Rapiscan 620DV, has the capability to bring these new technologies to airport checkpoints in the near future.

\section*{Acknowledgment}
This research has been funded by the Transportation Security Administration (TSA) under Contract
\#HSTS04-16-C-CT7020.
The authors would like to thank the TSA, specifically Armita Soroosh and Suriyun Whitehead, for their administrative support. 

\begin{acronym}[CCCCCCCC]
	\acro{AP}{Average Precision}
	\acro{AUC}{area-under-the-curve}
	\acro{CNN}{Convolutional Neural Network}
	\acro{$F_n$}{false negative}
	\acro{$F_p$}{false positive}
	\acro{GPU}{Graphical Processing Unit}
	\acro{IoU}{Intersection over Union}
	\acro{LAG}{liquids, aerosols, and gel}
	\acro{mAP}{mean Average Precision}
	\acro{PR}{Precision-recall}
	\acro{RGB}{Red-Green-Blue}
	\acro{RPN}{Region Proposal Network}
	\acro{SSD}{Single Shot MultiBox Detector}
	\acro{$T_p$}{true positive}
	\acro{TSA}{Transportation Security Administration}
	\acro{TSO}{Transportation Security Officer} 
\end{acronym}

{\small
\bibliographystyle{ieee_fullname}
\bibliography{refs}
}

\end{document}